\documentclass[letterpaper, 10 pt, journal, twoside]{ieeetran}
\usepackage[caption=false,font=normalsize,labelfont=sf,textfont=sf]{subfig}
\usepackage{textcomp}
\usepackage{stfloats}
\usepackage{url}
\usepackage{verbatim}
\usepackage{cite}
\usepackage{graphicx}
\usepackage{booktabs, multirow, hhline}
\usepackage[table]{xcolor}
\usepackage{amsmath,amsfonts}
\usepackage{tikz}
\usepackage{amssymb}
\usepackage{multirow}
\usepackage{booktabs}
\usepackage{makecell}
\usepackage{adjustbox}
\usepackage{tcolorbox}
\usepackage{listings}
\usepackage{array}
\usepackage{color, soul}
\usepackage{pifont}
\usepackage{tabularx}
\usepackage{algpseudocode}
\usepackage{xcolor}
\usepackage{diagbox}
\usepackage{hhline}
\usepackage{makecell}
 
\usepackage{amssymb}
\usepackage[ruled,vlined]{algorithm2e}

\begin{document}

\title{Breaking Déjà Vu: Independent Auditing of\\ Visual Place Recognition through Vision-Language Reasoning}

\author{Sania Waheed$^{1}$, Michael Milford$^{2}$, Sarvapali D. Ramchurn$^{1}$ and Shoaib Ehsan$^{1,3}$%
\thanks{$^{1}$Sania Waheed and Sarvapali D. Ramchurn are with the School of Electronics and Computer Science, University of Southampton, Southampton SO17 1BJ. 
        {\tt\footnotesize sw1m24@soton.ac.uk; sdr1@soton.ac.uk}}%
\thanks{$^{2}$M. Milford is with the School of Electrical Engineering and Computer Science, Queensland University of Technology, Brisbane, QLD 4000, Australia.
        {\tt\footnotesize michael.milford@qut.edu.au}}%
\thanks{$^{1,3}$Shoaib Ehsan is with the School of Electronics and Computer Science, University of Southampton, SO17 1BJ Southampton, U.K., and also with the School of Computer Science and Electronic Engineering, University of Essex, CO4 3SQ Colchester, U.K. {\tt\footnotesize (s.ehsan@soton.ac.uk)}}%
}



\maketitle

\begin{abstract}
Visual place recognition (VPR) is a key enabler of accurate localization and long-term autonomous navigation in robotics applications, such as loop closure detection for simultaneous localization and mapping (SLAM). However, real-world VPR deployment relies on selecting an image matching threshold that balances precision and recall. These thresholds are typically tuned using labeled validation data and fixed during deployment, making them unreliable under environmental changes where ground truth is unavailable. This is particularly problematic in safety-critical robotics, where accepting a false loop closure can corrupt the estimated trajectory and map. In this work, we introduce \textit{Visual Place Recognition Auditing}, an independent post-retrieval verification framework that leverages Vision-Language Models (VLMs) to assess retrieved matches by reasoning jointly over query and candidate images. Unlike conventional verification methods, our approach performs instance-level verification without architecture-specific or dataset-dependent threshold tuning, and without access to ground truth for operating point calibration. We evaluate our method on six benchmark datasets using five state-of-the-art VPR methods and four VLMs. Results show that VLM-based auditing is largely robust to prompt formulations, with the best-performing VLMs improving precision by 16\% on average while reducing false acceptance rates to 29\% and maintaining coverage at 79\%, at a cost of 1\% in recall@1.
\end{abstract}


\begin{IEEEkeywords}
Vision-Based Navigation, Localization, Mapping.
\end{IEEEkeywords}
\section{INTRODUCTION}\label{sec:introduction}
\IEEEPARstart{V}{PR} enables robots to localize themselves by matching current visual observations to a reference database. It is a core component of loop closure detection in SLAM systems~\cite{zaffar2021vpr, zhang2021visual, garg2021your}, where incorrect matches can corrupt the estimated trajectory and map~\cite{cadena2017past, lowry2015visual, zaffar2024estimation}, making higher verification precision a reasonable trade-off against inference time where real-time matching is not required \cite{zaffar2021vpr}. VPR systems require selecting an image matching threshold that determines whether a retrieved candidate is accepted as a valid match~\cite{schubert2021beyond, vysotska2025adaptive, rajani2026quantile}. This threshold is typically selected offline using precision-recall analysis on labeled validation data and fixed during deployment~\cite{schubert2021makes}. However, such thresholds may not generalize to unseen environments due to a number of appearance or environmental variations such as changes in viewpoint, illumination, weather, seasons, occlusions, and perceptual aliasing. In unknown environments where ground truth is unavailable, the system cannot determine whether its selected threshold provides an appropriate trade-off between false acceptance risk and coverage. This is particularly problematic in safety-critical applications where accepting an incorrect match can be significantly more damaging than missing a valid one.

\begin{figure} [t]
    \centering
    \vspace*{1mm}
    \includegraphics[width=0.9\columnwidth]{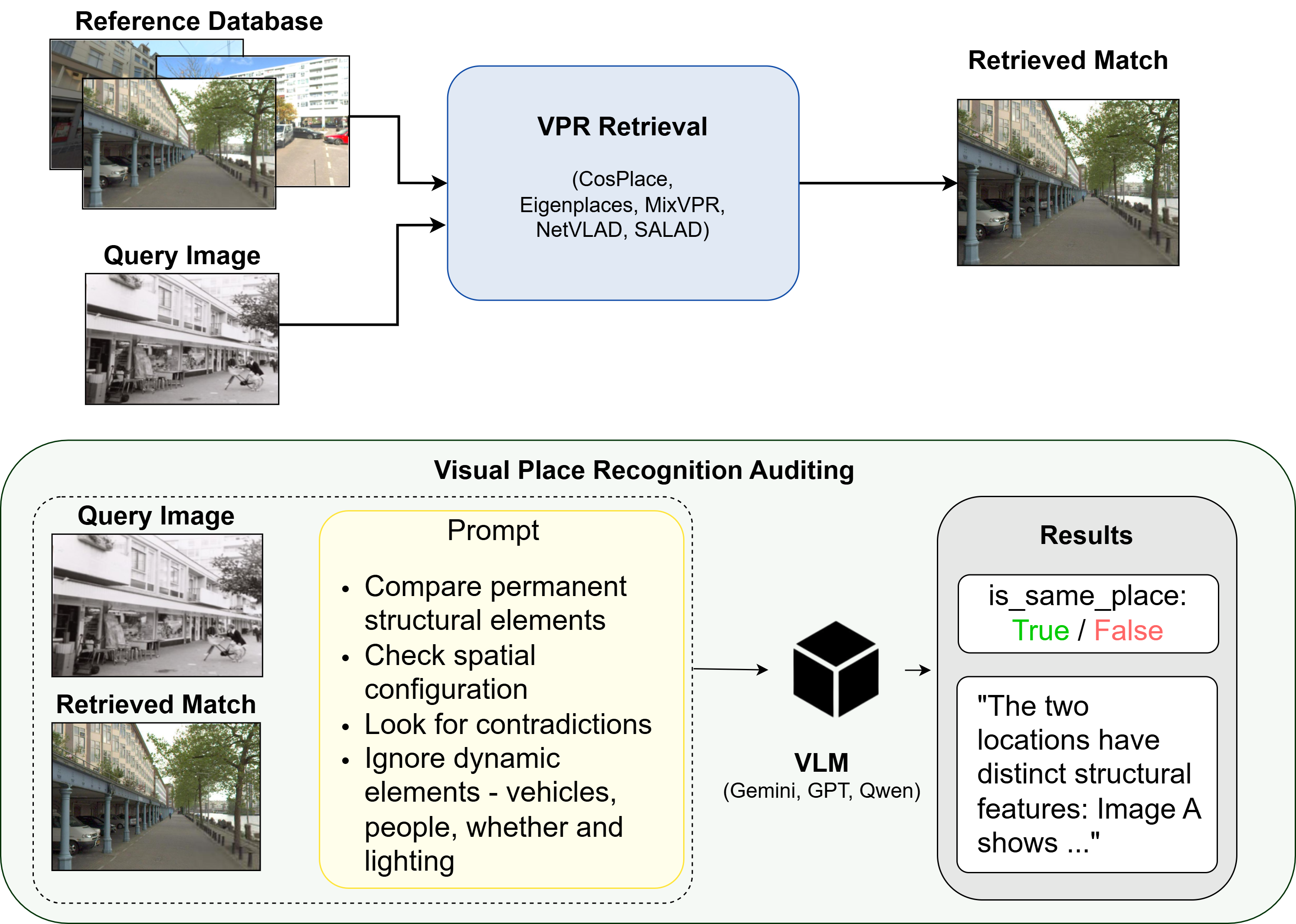}
    \caption{Proposed VPR Auditing framework: A standard VPR pipeline retrieves a candidate match for a given query image. The query and retrieved images are then passed to a VLM, which reasons over semantic and structural consistency to independently verify the retrieved match, outputting a binary accept/reject decision with supported reasoning.}
    \label{fig:AVPR}
\end{figure}
Existing approaches for mitigating false positives primarily focus on verification methods that estimate retrieved match reliability using geometric consistency or signals derived from image embeddings or similarity scores produced by the underlying VPR system~\cite{piasco2018survey, hausler2021unsupervised, warburg2021bayesian, cai2022stun}, assuming that match correctness can be inferred from these representations. In both cases, these methods require selecting appropriate operating point thresholds to convert verification scores into final decisions. 

To address these limitations, we introduce \textit{Visual Place Recognition Auditing}, a post-retrieval verification framework that independently assesses retrieved VPR matches. Instead of relying on thresholds selected using ground truth, VPR Auditing performs instance-level verification by directly evaluating whether a query and retrieved image represent the same physical location. We leverage VLMs to reason jointly over visual content and semantic context, providing a verification signal independent of the retrieval feature space, VPR architecture, and deployment environment. This allows zero-shot verification without prior knowledge of the environment or access to ground truth. We evaluate our approach across six benchmark datasets, five state-of-the-art VPR methods, and four VLMs. With the best-performing VLMs, our framework improves precision by 16\% on average over baseline VPR retrieval, while reducing false acceptance rates (FAR) to 29\% and maintaining coverage at 79\%, at a cost of 1\% in recall@1. We further conduct a prompt sensitivity analysis, which shows the overall robustness of the proposed framework to prompt formulation. Our main contributions are as follows:


\begin{itemize}
    \item We introduce \textit{Visual Place Recognition Auditing}, a post-retrieval verification framework that removes dependence on thresholds selected using ground truth.
    
    \item We demonstrate that VLMs provide an architecture-independent verification mechanism that rejects false positives while maintaining high precision and coverage across diverse VPR methods and environments.
    
    \item We show that AUC-PR masks poor false-positive detection and propose the precision, coverage, and FAR triplet with risk-coverage analysis for evaluation.
\end{itemize}

The remainder of this paper is organized as follows. Section~\ref{sec:related_works} reviews related work in VPR, match verification, operating point selection and VLMs. Section~\ref{sec:method} presents the proposed VPR auditing framework and evaluation metrics. Section~\ref{sec:experiment} describes the experimental details of datasets, VPR methods, VLMs and the auditing protocol followed by a detailed analysis and discussion of results in Section~\ref{sec:results}. Finally, Section~\ref{sec:conclusion} concludes the paper.

\section{RELATED WORK}\label{sec:related_works}
VPR systems typically retrieve candidate matches using global image descriptors, with additional verification mechanisms often applied to assess retrieval reliability. Although recent advances in loss functions~\cite{leyva2021generalized, revaud2019learning}, 
training data~\cite{ali2022gsv, berton2022rethinking}, architectures~\cite{wang2022transvpr, 
zhang2021vector}, and aggregation strategies~\cite{ali2023mixvpr, arandjelovic2016netvlad} 
have significantly improved retrieval accuracy, reliably identifying correct matches remains challenging, particularly in safety-critical systems where false loop closures can corrupt downstream SLAM pipelines~\cite{zaffar2021vpr, cadena2017past}. This has motivated verification approaches that aim to reject unreliable predictions and improve localization robustness~\cite{zaffar2024estimation, miller2026through}. Existing methods can be broadly categorized into retrieval-based uncertainty estimation(RUE), data-driven uncertainty estimation(DUE), spatial uncertainty estimation(SUE), and geometric verification(GV).

RUE methods estimate confidence directly from descriptor similarity, using measures such as nearest-neighbour distance or distance ratios~\cite{piasco2018survey, hausler2021unsupervised}. While computationally efficient, they depend strongly on the learned embedding space and may become overconfident under perceptual aliasing~\cite{zaffar2024estimation}. DUE methods instead learn uncertainty-aware representations by modeling descriptors as distributions with associated uncertainty estimates, such as Bayesian Triplet Loss (BTL)~\cite{warburg2021bayesian} and STUN~\cite{cai2022stun}; however, they require specialized training and often generalize poorly across domains~\cite{miller2026through}. SUE estimates uncertainty from the spatial distribution of top-K retrieved poses~\cite{zaffar2024estimation}, but requires geo-tagged reference images and assumes comparable spatial distributions between query and reference data. GV methods verify matches through local feature correspondences and geometric consistency using approaches such as SIFT~\cite{Lowe2004DistinctiveIF}, DELF~\cite{noh2017large}, and SuperPoint~\cite{detone2018superpoint}. However, geometric verification is computationally expensive and remains vulnerable to repetitive structures and geometric burstiness, where incorrect locations may still produce consistent local matches~\cite{sattler2016large, zaffar2024estimation}. Furthermore, all these methods require selecting a threshold to convert confidence scores into final accept/reject decisions. These thresholds control the precision-recall trade-off and are commonly selected offline using ground truth data from a specific deployment environment~\cite{rajani2026quantile}. Fixed thresholds often fail to generalize under environmental changes. Recent works have explored adaptive threshold selection~\cite{rajani2026quantile, mount2020unsupervised}, but still rely on environment-specific assumptions, calibration data, or prior knowledge of deployment conditions.


VLMs have recently emerged as powerful tools leveraging large-scale vision-language pretraining to enable scene understanding and multimodal reasoning for robotic perception, navigation, and localization~\cite{wu2022im2city, waheed2025vlm, waheed2025image, ahn2022can, yu2023language}. Structured prompting strategies have further been explored to decrease prompt sensitivity and improve the consistency and reliability of VLM reasoning by guiding model outputs~\cite{waheed2025image}. Recent efforts have also investigated efficient VLM deployment through distillation, enabling lightweight multimodal reasoning for resource-constrained robotic platforms~\cite{baumann2026localnav}. In the context of localization, VLMs have been explored for image-based geo-localization~\cite{waheed2025vlm, waheed2025image}; however, these approaches primarily focus on direct localization rather than verifying retrieved candidates. To the best of our knowledge, the use of VLMs as an independent post-retrieval verification mechanism for VPR remains unexplored.

\section{METHODOLOGY}\label{sec:method}
We propose \textit{Visual Place Recognition Auditing}, a post-retrieval framework that performs instance-level verification of each retrieved match without relying on thresholds selected using ground truth. The framework integrates with any existing VPR pipeline as an independent layer, requiring no retraining or modification of the underlying model.

\subsection{Problem Formulation}
Let $\mathcal{D} = \{d_1, \ldots, d_N\}$ denote a reference database and $f$ denote a VPR retrieval function. Given a query image $q$, $f$ returns a top-1 candidate $r = f(q;\,\mathcal{D})$, which is typically accepted directly as a localization estimate. While this is sufficient for many applications, safety-critical systems require additional assurance since an incorrect loop closure can corrupt the map. In such settings, conventional VPR systems accept a candidate using a threshold on confidence scores, calibrated offline on labeled data from a specific environment. To address this limitation, we introduce a post-retrieval verifier which independently assesses whether the retrieved candidate represents the same physical location:
\[
V:(q,r) \mapsto \{\textit{accept},\,\textit{reject}\}.
\]
Existing verification methods typically estimate match correctness from the same representation used for retrieval. Formally, if the retrieval network computes descriptors: $\mathbf{z}_q = f(q)$ and $\mathbf{z}_r = f(r)$, then confidence estimators operate on $g(\mathbf{z}_q,\mathbf{z}_r)$ where $g$ represents descriptor distance or uncertainty estimates. Consequently, their decision is limited to the information retained by the retrieval representation. Since global descriptors compress observations into embeddings optimized for retrieval rather than verification, discarded scene information cannot be recovered downstream. In contrast, a VLM reasons directly over the original image pair $(q,r)$, allowing it to exploit semantic, structural, and relational cues that may not be preserved in descriptor space. Geometric verification similarly operates on the image pair, but requires selecting a threshold on local feature correspondences for each environment. Independence from the retrieval representation and the absence of a numerical decision threshold together enable architecture-agnostic verification without environment-specific calibration.

\subsection{VPR Verification using Vision-Language Models}
We implement $V$ using a Vision-Language Model (VLM), which receives the query image $q$ and the retrieved image $r$ and is prompted to determine whether both images depict the same physical place. The VLM is used zero-shot, since training a verifier would require labeled data from the deployment environment and reintroduce the dependency the framework is intended to remove. As illustrated in Fig.~\ref{fig:AVPR}, the VLM performs joint reasoning over the image pair, evaluating semantic and structural consistency, specifically including scene layout, architectural elements, street geometry, and distinctive landmarks.

The VLM is prompted using a conservative verification strategy, where it is instructed to assume that the images represent different locations unless strong evidence supports a match. This decision strategy reflects the requirements of safety-critical robotics where accepting an incorrect loop closure can corrupt the map, whereas rejecting a valid match merely misses a loop closure that can be recovered later. The prompt therefore expresses a decision policy, but unlike threshold selection it is applied unchanged across all environments and requires no ground-truth calibration. The sensitivity of the framework to this choice is evaluated in Section~\ref{sec:prompt_sensitivity}. The VLM returns a structured response containing a binary accept/reject decision and an explanation.


\begin{table}[t]
\vspace{2mm}
\centering
\caption{Summary of datasets used for evaluation.}
\label{tab:datasets}
\setlength{\tabcolsep}{3pt}
\renewcommand{\arraystretch}{1.0}
\footnotesize
\begin{tabularx}{\columnwidth}
{>{\raggedright\arraybackslash}p{0.20\columnwidth}
 >{\raggedright\arraybackslash}X
 r r c}
\hline
\textbf{Dataset} & \textbf{Challenge} & \textbf{\#DB} & \textbf{\#Q} & \textbf{Eval.} \\
\hline
Pitts250k \cite{arandjelovic2016netvlad} &
Repetitive urban structures & 83,952 & 8,280 & $\leq$25\,m \\
StLucia \cite{milford2008mapping} &
Weakly distinctive suburbs & 1,464 & 1,464 & $\leq$25\,m \\
Tokyo24/7 \cite{torii201524} &
Day--night and weather & 75,984 & 315 & $\leq$25\,m \\
SanFran \cite{berton2022rethinking} &
Illumination and viewpoint & 27,191 & 1,000 & $\leq$25\,m \\
Nordland \cite{sunderhauf2013we} &
Severe cross-season change & 3,066 & 3,066 & 1:1 \\
AmsterTime \cite{yildiz2022amstertime} &
Long-term structural change & 1,231 & 1,231 & 1:1 \\
\hline
\end{tabularx}
\end{table}

\subsection{Proposed Evaluation Framework}
Match verification methods are commonly evaluated using AUC-PR values. We argue that AUC-PR alone is insufficient for evaluating post-retrieval verification systems for two reasons. First, AUC-PR aggregates performance across all possible decision thresholds, whereas conventional verification systems operate at a single selected operating point during deployment. Second, AUC-PR does not directly measure whether a verifier can reject false positives. When baseline precision $p_0$ is already high, false positives represent only a small fraction of all retrieved matches. Consequently, a verifier that accepts nearly all candidates has a PR curve approximately flat at $p_0$, yielding a high AUC-PR value ($\approx p_0$) despite providing limited false-positive detection.


The false acceptance rate (FAR) directly captures this failure mode by measuring the fraction of false matches that are accepted. A verifier may therefore achieve high AUC-PR while having FAR close to 1.0, indicating that it provides little protection against false loop closures. To better characterize system reliability, we evaluate verification methods using the proposed precision, coverage, and FAR triplet together with risk-coverage curves.

\subsubsection{Precision-Coverage-FAR Triplet}
Let $\text{TP}_\tau$, $\text{FP}_\tau$, $\text{TN}_\tau$, and $\text{FN}_\tau$ denote true positives, false positives, true negatives, and false negatives at a fixed operating threshold, $\tau$, and let $N$ be the total number of query-candidate pairs. We define:
\begin{align}
    \mathcal{P}_\tau   &= \text{TP}_\tau / (\text{TP}_\tau + \text{FP}_\tau), \label{eq:precision} \\[3pt]
    \mathcal{C}_\tau   &= (\text{TP}_\tau + \text{FP}_\tau) / N, \label{eq:coverage} \\[3pt]
    \text{FAR}_\tau    &= \text{FP}_\tau / (\text{FP}_\tau + \text{TN}_\tau) \label{eq:far}
\end{align}
Precision measures the correctness of accepted matches. Coverage measures the fraction of queries for which the system provides an accepted match, preventing methods from achieving high precision by rejecting almost all candidates. FAR measures the proportion of false positives that are accepted, penalizing methods that achieve high coverage by accepting matches indiscriminately. Together, precision, coverage, and FAR characterize both safety (precision and FAR) and availability (coverage). We note that $Recall@1 = TP_\tau/N$, where $N$ is the total number of queries. For the unaudited system all candidates are accepted, so $C_\tau = 1$ and Recall@1 coincides with precision. Since a post-hoc verifier can only remove candidates, it cannot increase Recall@1 over the unaudited system.

To enable comparison across methods, we define a composite score using the harmonic mean of precision $\mathcal{P}$, coverage $\mathcal{C}$, and $(1-\text{FAR})$, which penalizes poor performance in any dimension:
\begin{equation}\label{eq:score}
\mathcal{S} = \frac{3}{\frac{1}{\mathcal{P}_\tau} + \frac{1}{\mathcal{C}_\tau} + \frac{1}{1 - \text{FAR}_\tau}}.
\end{equation}


\subsubsection{Risk-Coverage Curves}
Risk-coverage curves are constructed by varying the decision threshold $\tau$ and plotting risk $\mathcal{R}_\tau$ against coverage $\mathcal{C}_\tau$, where:
\begin{align}
    \mathcal{R} = 1-\mathcal{P}_\tau
\end{align}
A lower risk-coverage curve indicates lower false acceptance rates for a given level of system availability. 

This framework exposes failure modes missed by AUC-PR, distinguishing methods that genuinely reject false loop closures from those that achieve high precision through indiscriminate acceptance or rejection.

\section{EXPERIMENTAL SETUP}\label{sec:experiment}
This section outlines the datasets, VPR methods, VLM configurations, auditing protocol, and evaluation metrics used throughout the experiments.

\subsection{Datasets}\label{sec:experiment-datasets}
We evaluate the proposed auditing framework on six widely used VPR benchmarks covering diverse deployment challenges, including perceptual aliasing, appearance variation, and long-term distribution shift. Table~\ref{tab:datasets} summarizes the datasets, their challenges, and evaluation protocols.

\subsection{Visual Place Recognition Methods}
To evaluate generalization across retrieval pipelines, we consider five VPR architectures spanning global descriptor learning, aggregation-based methods, and transformer-based approaches: EigenPlaces~\cite{berton2023eigenplacestrainingviewpointrobust}, MixVPR~\cite{ali2023mixvpr}, CosPlace~\cite{berton2022rethinking}, NetVLAD~\cite{arandjelovic2016netvlad}, and SALAD~\cite{Izquierdo_CVPR_2024_SALAD}. For each method, we follow the standard evaluation protocol and use the top-1 retrieved image as the candidate match.

\subsection{Vision-Language Models}\label{sec:experiment-vlms}
We evaluate four VLMs spanning proprietary and open-weight families, covering different model scales and training paradigms. All models use a unified inference pipeline to ensure comparability and avoid confounding effects from prompt design or implementation. \textbf{Closed-source models.} We include OpenAI GPT-4.1 (gpt-4.1) and Google's Gemini-3-Flash-Preview (gemini-3-flash-preview). \textbf{Open-weight models.} We evaluate Alibaba's Qwen2.5-VL family, specifically the 7B-Instruct and 32B-Instruct variants, using the publicly available Hugging Face checkpoints and inference code. Qwen models were run locally on NVIDIA H200 GPUs, while GPT-4.1 and Gemini were accessed through their provider APIs. Per-query inference time for all four models is reported in Table~\ref{tab:operating_point}. All experiments use greedy decoding. Since GPT-4.1 and Gemini-3-Flash-Preview are provider-managed model aliases rather than fixed snapshots, their behavior may change over time; all experiments were conducted in March 2026. Results for Qwen models are reproducible from the cited checkpoints.

\subsection{Prompting and Decision Protocol}\label{sec:experiment-prompting}
The auditing decision is obtained by prompting the VLM with the query and candidate image. The prompt is designed to encourage structured reasoning and discourage reliance on superficial visual similarity. Specifically, the VLM is instructed to: (i) identify permanent structural elements in both images, (ii) compare their spatial configuration, and (iii) actively search for inconsistencies. Dynamic elements such as vehicles, pedestrians, lighting, and weather conditions are explicitly ignored. The prompt does not provide dataset-specific calibration information, the VLM produces an instance-level verification decision from the semantic and structural evidence in each image pair.

The same prompt is used for all datasets, backbones and models in the main experiments, and is denoted P3 in what follows. To assess sensitivity to this design choice, we evaluate three further variants: permissive (P1) and neutral (P2) loosen the decision rule, while P4 expresses the same conservative policy as P3 in different wording, with a semantic similarity of 0.90 and 11\% shared phrases.



The VLM returns a structured JSON response; the final auditing decision is derived directly from the binary \texttt{is\_same\_place} field, while the \texttt{reasoning} field provides interpretable justification for qualitative analysis. If the VLM refuses to answer or produces an invalid response, the retrieved match is retained and treated as accepted.

\subsection{Evaluation Metrics}\label{sec:experiment-evaluation}
We evaluate all verification methods using the precision-coverage-FAR triplet and risk-coverage curves. For VLM-based auditing, no threshold calibration, validation data, or environment-specific tuning is required. For baselines producing confidence scores, we evaluate two operating points: a fixed threshold of 0.5 after normalization, and a threshold that maximizes the harmonic mean of precision and coverage using ground-truth labels as an optimistic estimate achievable through manual calibration.
\section{RESULTS}\label{sec:results}
We evaluate VLM auditing from four perspectives: the effect of auditing on precision and Recall@1 compared with VPR and verification baselines (Sec.~\ref{sec:results-recall1}), safety--availability trade-offs using precision, coverage, and FAR (Sec.~\ref{sec:results-triplet}, \ref{sec:results-riskcoverage}), the limitations of AUC-PR as a verification metric (Sec.~\ref{sec:results-aucpr}), and the sensitivity of the framework to prompt formulation (Sec.~\ref{sec:prompt_sensitivity}).

\begin{table}[t]
\vspace{2mm}
\centering
\caption{Precision comparison of VPR methods with and without VLM verification across six benchmark datasets and four VLMs.}
\label{tab:vpr_vlm_results}
\scriptsize
\setlength{\tabcolsep}{1.5pt}
\renewcommand{\arraystretch}{0.85}
\resizebox{\columnwidth}{!}{
\begin{tabular}{l|l|cc|cc|cc|cc|cc|cc}
\toprule
\multirow{2}{*}{\makecell{\textbf{VPR}\\\textbf{Method}}} & 
\multirow{2}{*}{\textbf{VLM}}
& \multicolumn{2}{c|}{\textbf{Amster}}
& \multicolumn{2}{c|}{\textbf{Pitts}}
& \multicolumn{2}{c|}{\textbf{SF}}
& \multicolumn{2}{c|}{\textbf{StL}}
& \multicolumn{2}{c|}{\textbf{Tokyo}}
& \multicolumn{2}{c}{\textbf{Nord}} \\
\cmidrule(lr){3-14}
&
& \textbf{Base} & \textbf{+VLM}
& \textbf{Base} & \textbf{+VLM}
& \textbf{Base} & \textbf{+VLM}
& \textbf{Base} & \textbf{+VLM}
& \textbf{Base} & \textbf{+VLM}
& \textbf{Base} & \textbf{+VLM} \\
\midrule
\multirow{4}{*}{CosPlace}
& Gemini
& \multirow{4}{*}{0.46} & 0.73
& \multirow{4}{*}{0.92} & 0.95
& \multirow{4}{*}{0.75} & 0.96
& \multirow{4}{*}{0.99} & 1.00
& \multirow{4}{*}{0.90} & 0.98
& \multirow{4}{*}{0.70} & 0.91 \\
& GPT      
&  & 0.72 &  & 0.95 &  & 0.94 &  & 1.00 &  & 0.97 &  & 0.93 \\
& Qwen-7b  
&  & 0.74 &  & 0.97 &  & 0.97 &  & 1.00 &  & 0.99 &  & 1.00 \\
& Qwen-32b 
&  & 0.62 &  & 0.95 &  & 0.94 &  & 0.99 &  & 0.98 &  & 0.99 \\
\midrule
\multirow{4}{*}{EigenPlaces}
& Gemini
& \multirow{4}{*}{0.46} & 0.71
& \multirow{4}{*}{0.93} & 0.96
& \multirow{4}{*}{0.82} & 0.97
& \multirow{4}{*}{1.00} & 1.00
& \multirow{4}{*}{0.89} & 0.98
& \multirow{4}{*}{0.72} & 0.92 \\
& GPT      
&  & 0.71 &  & 0.96 &  & 0.94 &  & 1.00 &  & 0.96 &  & 0.94 \\
& Qwen-7b  
&  & 0.73 &  & 0.97 &  & 0.97 &  & 1.00 &  & 0.99 &  & 1.00 \\
& Qwen-32b 
&  & 0.62 &  & 0.96 &  & 0.95 &  & 1.00 &  & 0.97 &  & 1.00 \\
\midrule
\multirow{4}{*}{MixVPR}
& Gemini
& \multirow{4}{*}{0.36} & 0.71
& \multirow{4}{*}{0.93} & 0.96
& \multirow{4}{*}{0.72} & 0.96
& \multirow{4}{*}{0.99} & 1.00
& \multirow{4}{*}{0.79} & 0.96
& \multirow{4}{*}{0.69} & 0.92 \\
& GPT      
&  & 0.71 &  & 0.96 &  & 0.95 &  & 1.00 &  & 0.94 &  & 0.94 \\
& Qwen-7b  
&  & 0.75 &  & 0.97 &  & 0.98 &  & 1.00 &  & 0.99 &  & 1.00 \\
& Qwen-32b 
&  & 0.61 &  & 0.96 &  & 0.94 &  & 1.00 &  & 0.96 &  & 0.98 \\
\midrule
\multirow{4}{*}{NetVLAD}
& Gemini
& \multirow{4}{*}{0.16} & 0.62
& \multirow{4}{*}{0.86} & 0.92
& \multirow{4}{*}{0.57} & 0.95
& \multirow{4}{*}{0.64} & 0.94
& \multirow{4}{*}{0.70} & 0.96
& \multirow{4}{*}{0.15} & 0.72 \\
& GPT      
&  & 0.69 &  & 0.91 &  & 0.92 &  & 0.94 &  & 0.97 &  & 0.77 \\
& Qwen-7b  
&  & 0.74 &  & 0.94 &  & 0.95 &  & 0.93 &  & 0.98 &  & 0.97 \\
& Qwen-32b 
&  & 0.57 &  & 0.92 &  & 0.92 &  & 0.84 &  & 0.98 &  & 0.98 \\
\midrule
\multirow{4}{*}{SALAD}
& Gemini
& \multirow{4}{*}{0.61} & 0.78
& \multirow{4}{*}{0.95} & 0.96
& \multirow{4}{*}{0.87} & 0.98
& \multirow{4}{*}{1.00} & 1.00
& \multirow{4}{*}{0.97} & 1.00
& \multirow{4}{*}{0.89} & 0.95 \\
& GPT      
&  & 0.76 &  & 0.96 &  & 0.96 &  & 1.00 &  & 0.99 &  & 0.96 \\
& Qwen-7b  
&  & 0.74 &  & 0.98 &  & 0.99 &  & 1.00 &  & 1.00 &  & 1.00 \\
& Qwen-32b 
&  & 0.70 &  & 0.97 &  & 0.97 &  & 1.00 &  & 1.00 &  & 1.00 \\
\bottomrule
\end{tabular}
}
\end{table}

\begin{figure*}[ht]
\vspace{2mm}
    \centering
    \includegraphics[width=0.85\textwidth]{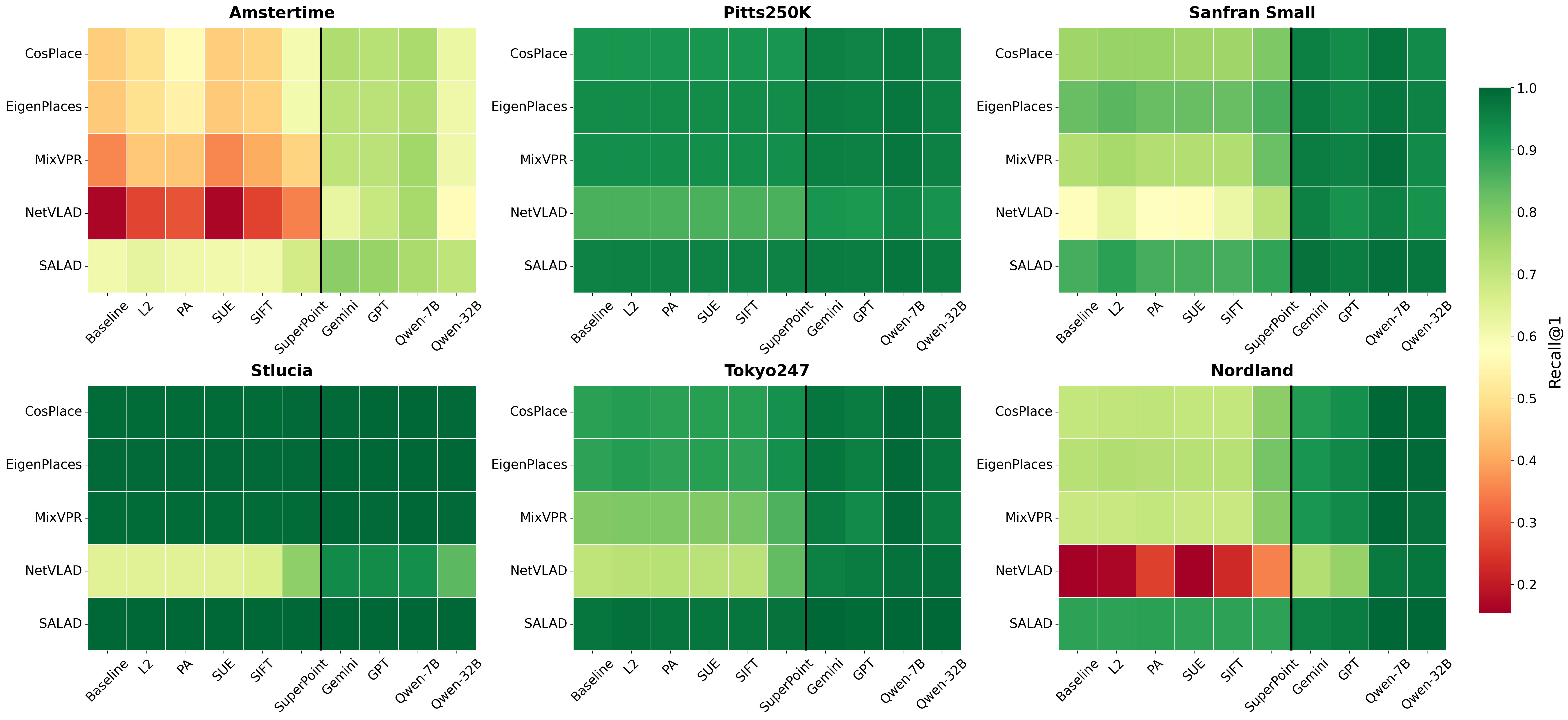}
    \caption{Precision after filtering for every combination of dataset, VPR backbone, and verification method, including the unfiltered baseline (leftmost column) for reference. Conventional verification methods report precision at the selected threshold $\tau$, while VLMs use a fixed binary accept/reject decision.}
    \label{fig:precision}
\end{figure*}

\begin{table*}[t]
\centering
\caption{Safety--availability comparison per dataset, averaged across VPR backbones, under two threshold-selection strategies. (a) $\tau$ at normalized confidence 0.5, (b) $\tau$ maximising the harmonic mean of precision and coverage. VLMs use a fixed zero-shot accept/reject decision. $\mathcal{P}$ precision, $\mathcal{C}$ coverage, FAR false acceptance rate, $\mathcal{S}$ aggregate score (Eq.~\ref{eq:score}). Inference time per query is from~\cite{miller2026through} for the uncertainty metrics (NVIDIA A100) and measured here for the VLMs (NVIDIA H200 for local models, end-to-end API round-trip for hosted models). Best and second-best are \textbf{bold} and \underline{underlined}.}

\label{tab:operating_point}
\setlength{\tabcolsep}{3.5pt}
\renewcommand{\arraystretch}{1.15}
\setlength{\arrayrulewidth}{0.8pt}

\resizebox{\textwidth}{!}{
\begin{tabular}{|l|ccc|c|ccc|c|ccc|c|ccc|c|ccc|c|ccc|c|c|c|}
\hline

\multicolumn{1}{|c|}{\textbf{Datasets $\rightarrow$}}
& \multicolumn{4}{c|}{AmsterTime}
& \multicolumn{4}{c|}{Pitts250k}
& \multicolumn{4}{c|}{SanFran}
& \multicolumn{4}{c|}{StLucia}
& \multicolumn{4}{c|}{Tokyo247}
& \multicolumn{4}{c|}{Nordland}
& \multicolumn{1}{c|}{}
& \multicolumn{1}{c|}{} \\

\hline

\textbf{Method $\downarrow$}
& $\mathcal{P}$ & $\mathcal{C}$ & FAR & $\mathcal{S}$
& $\mathcal{P}$ & $\mathcal{C}$ & FAR & $\mathcal{S}$
& $\mathcal{P}$ & $\mathcal{C}$ & FAR & $\mathcal{S}$
& $\mathcal{P}$ & $\mathcal{C}$ & FAR & $\mathcal{S}$
& $\mathcal{P}$ & $\mathcal{C}$ & FAR & $\mathcal{S}$
& $\mathcal{P}$ & $\mathcal{C}$ & FAR & $\mathcal{S}$
& Avg. $\mathcal{S}$ & Time (s) \\

\hline

\multicolumn{27}{|c|}{(a) $\tau \geq0.5$ on normalized confidence scores (\textit{label-free threshold)}}\\

\hline

L2
& 0.83 & 0.06 & 0.02 & 0.16
& 0.99 & 0.11 & 0.05 & 0.28
& 0.98 & 0.38 & 0.02 & 0.64
& 0.99 & 0.58 & 0.00 & 0.80
& 0.99 & 0.28 & 0.01 & 0.54
& 0.82 & 0.25 & 0.05 & 0.48
& 0.48 & $<0.01$ \\

PA
& 0.99 & 0.03 & 0.00 & 0.09
& 1.00 & 0.04 & 0.00 & 0.10
& 1.00 & 0.05 & 0.00 & 0.13
& 1.00 & 0.09 & 0.00 & 0.24
& 1.00 & 0.12 & 0.00 & 0.30
& 1.00 & 0.07 & 0.00 & 0.19
& 0.17 & $<0.01$ \\

SIFT
& 0.49 & 0.65 & 0.57 & 0.51
& 0.94 & 0.92 & 0.72 & 0.52
& 0.80 & 0.86 & 0.68 & 0.55
& 0.95 & 0.91 & 0.54 & 0.69
& 0.88 & 0.90 & 0.64 & 0.59
& 0.73 & 0.63 & 0.39 & 0.65
& 0.59 & 0.14 \\

SUE
& 0.40 & 0.07 & 0.08 & 0.18
& 0.92 & 0.97 & 0.94 & 0.16
& 0.95 & 0.28 & 0.06 & 0.53
& 0.92 & 0.93 & 0.91 & 0.23
& 0.87 & 0.87 & 0.72 & 0.51
& 0.56 & 0.14 & 0.19 & 0.30
& 0.32 & $<0.01$ \\

SuperPoint
& 0.84 & 0.04 & 0.01 & 0.10
& 1.00 & 0.02 & 0.00 & 0.07
& 1.00 & 0.09 & 0.00 & 0.23
& 1.00 & 0.22 & 0.00 & 0.46
& 1.00 & 0.30 & 0.00 & 0.56
& 1.00 & 0.02 & 0.00 & 0.05
& 0.25 & 0.08 \\

\hline

\multicolumn{27}{|c|}{(b) $\tau=$ best harmonic mean\textit{(ground-truth dependent)}}\\

\hline

L2
& 0.47 & 0.79 & 0.72 & 0.43
& 0.92 & 1.00 & 0.99 & 0.02
& 0.77 & 0.96 & 0.85 & 0.33
& 0.92 & 1.00 & 0.88 & 0.28
& 0.86 & 0.99 & 0.88 & 0.29
& 0.63 & 0.94 & 0.93 & 0.18
& 0.26 & $<0.01$ \\

PA
& 0.49 & 0.74 & 0.66 & 0.48
& 0.92 & 1.00 & 1.00 & 0.00
& 0.75 & 0.99 & 0.98 & 0.05
& 0.92 & 1.00 & 1.00 & 0.00
& 0.86 & 1.00 & 0.98 & 0.05
& 0.65 & 0.89 & 0.86 & 0.31
& 0.15 & $<0.01$ \\

SIFT
& 0.44 & 0.86 & 0.82 & 0.33
& 0.92 & 1.00 & 1.00 & 0.00
& 0.76 & 0.98 & 0.95 & 0.12
& 0.93 & 0.99 & 0.99 & 0.04
& 0.86 & 0.99 & 0.97 & 0.08
& 0.64 & 0.91 & 0.90 & 0.23
& 0.13 & 0.14 \\

SUE
& 0.41 & 1.00 & 1.00 & 0.00
& 0.92 & 1.00 & 1.00 & 0.00
& 0.75 & 1.00 & 1.00 & 0.00
& 0.92 & 1.00 & 1.00 & 0.00
& 0.86 & 1.00 & 0.97 & 0.08
& 0.63 & 1.00 & 1.00 & 0.00
& 0.01 & $<0.01$ \\

SuperPoint
& 0.54 & 0.72 & 0.57 & 0.54
& 0.92 & 1.00 & 1.00 & 0.00
& 0.82 & 0.91 & 0.69 & 0.54
& 0.95 & 0.96 & 0.66 & 0.59
& 0.90 & 0.94 & 0.66 & 0.59
& 0.72 & 0.81 & 0.63 & 0.56
& 0.47 & 0.08 \\

\hline

\multicolumn{27}{|c|}{Vision--Language Models \textit{(Zero-shot decision)}}\\

\hline

Gemini
& 0.71 & 0.55 & 0.28 & \textbf{0.65}
& 0.95 & 0.95 & 0.59 & \underline{0.66}
& 0.96 & 0.77 & 0.12 & \textbf{0.86}
& 0.99 & 0.93 & 0.29 & \textbf{0.86}
& 0.97 & 0.87 & 0.16 & \textbf{0.89}
& 0.88 & 0.68 & 0.23 & \underline{0.77}
& \textbf{0.78} & 6.93 \\

GPT-4.1
& 0.72 & 0.54 & 0.28 & \underline{0.65}
& 0.95 & 0.96 & 0.65 & 0.61
& 0.94 & 0.79 & 0.20 & \underline{0.84}
& 0.99 & 0.93 & 0.30 & \underline{0.85}
& 0.96 & 0.88 & 0.27 & 0.85
& 0.91 & 0.64 & 0.16 & \textbf{0.78}
& \underline{0.76} & 1.85 \\

Qwen-7B
& 0.74 & 0.18 & 0.09 & 0.37
& 0.97 & 0.82 & 0.36 & \textbf{0.78}
& 0.97 & 0.60 & 0.06 & 0.80
& 0.98 & 0.80 & 0.28 & 0.82
& 0.99 & 0.55 & 0.02 & 0.78
& 0.99 & 0.01 & 0.00 & 0.03
& 0.60 & 0.52 \\

Qwen-32B
& 0.62 & 0.52 & 0.36 & 0.59
& 0.95 & 0.94 & 0.61 & 0.64
& 0.94 & 0.70 & 0.16 & 0.82
& 0.97 & 0.94 & 0.38 & 0.81
& 0.98 & 0.76 & 0.10 & \underline{0.87}
& 0.99 & 0.13 & 0.00 & 0.30
& 0.67 & 2.51 \\

\hline
\end{tabular}
}
\end{table*}

\begin{figure*}[t]
\vspace{2mm}
    \centering
    \includegraphics[width=0.85\textwidth]{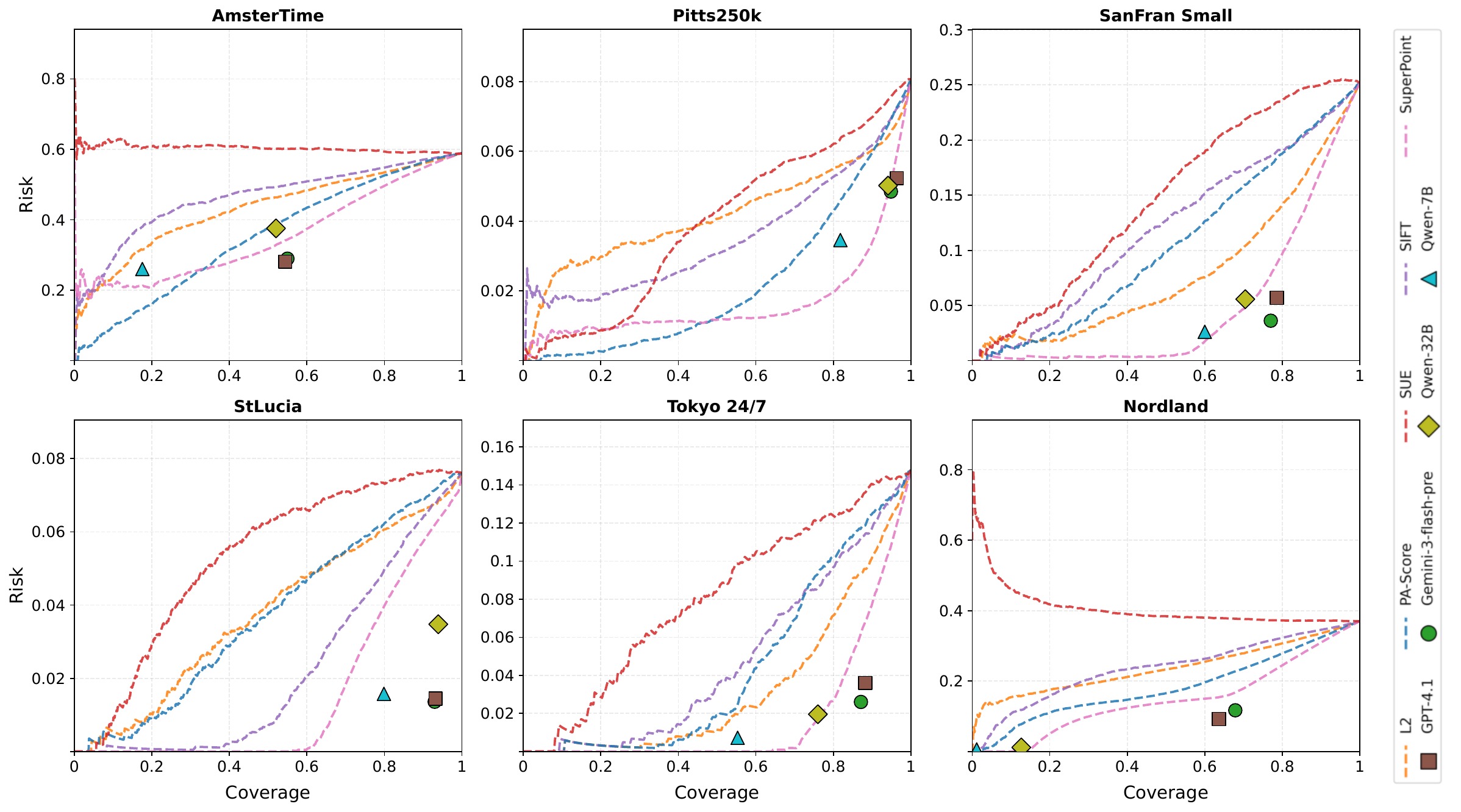}
    \caption{Risk--coverage curves per dataset, averaged across VPR backbones. Lines show threshold-swept verification  methods; markers show VLM operating points. Lower risk at higher coverage indicates better verification performance.}
    \label{fig:risk_coverage}
\end{figure*}

\begin{figure*}[t]
    \centering
    \includegraphics[width=0.9\textwidth]{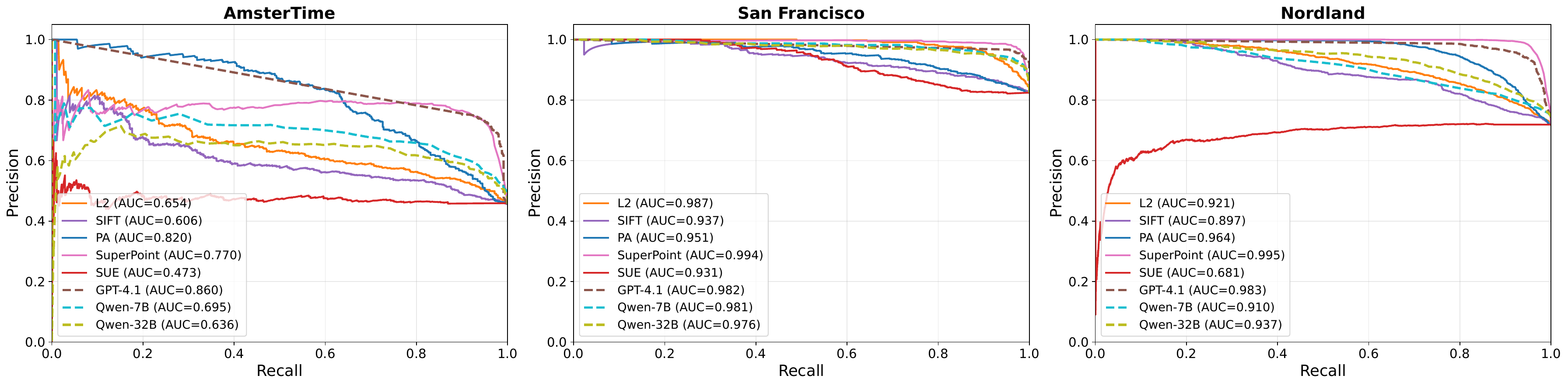}
    \caption{Precision--recall curves for three representative datasets (AmsterTime, SanFran Small, Nordland), using EigenPlaces as the VPR backbone. Uncertainty metrics are swept across confidence thresholds; VLM curves (GPT-4.1, Qwen-7B, Qwen-32B) use token-level log-probability of ``true'' as a confidence score. Gemini is omitted as its API does not expose token-level log-probabilities. Similar AUC-PR values can correspond to substantially different verification behaviour, motivating operating-point metrics such as precision, coverage, and FAR.}
    \label{fig:pr_curves}
\end{figure*}

\subsection{Precision and Recall@1 Comparison} \label{sec:results-recall1}
We report results across six benchmark datasets and five VPR backbones, using four VLMs (GPT-4.1, Gemini-3-Flash-Preview, Qwen-32b and Qwen-7b) and five baseline verification methods (L2 distance\cite{piasco2018survey}, PA score\cite{hausler2021unsupervised}, SIFT\cite{Lowe2004DistinctiveIF}, SuperPoint\cite{detone2018superpoint}, and SUE\cite{zaffar2024estimation}). Table~\ref{tab:vpr_vlm_results} compares precision for baseline VPR and VLM auditing. On average VLMs improve precision by 17\%, with the largest gains on challenging datasets: AmsterTime improves from 0.16--0.61 to 0.62--0.78, while Nordland improves from 0.15--0.89 to 0.91--1.00. Since the auditor can only remove candidates, this gain is obtained at the cost of Recall@1. Averaged over all backbones and datasets, Gemini and GPT-4.1 raise precision from 0.747 to 0.912 and 0.911 respectively, while reducing Recall@1 to 0.739 and 0.736, a cost of 0.8\% and 1.1\%. The smaller models reach similar precision, with Recall@1 falling to 0.612 for Qwen-32B and 0.476 for Qwen-7B.

Fig.~\ref{fig:precision} shows precision for each verifier at its selected threshold. VLMs consistently outperform all verification baselines. SuperPoint, the strongest conventional baseline, improves precision by 0--13\% over the VPR baselines (0\% on Pitts250k, 13\% on AmsterTime), whereas VLM verification provides 5--33\% improvements on the same datasets (5\% on Pitts250k, 33\% on AmsterTime).

AmsterTime is the most challenging dataset, whereas Pitts250k and StLucia already average 0.92 precision. Datasets with already-high baseline precision provide little room for verification method analysis based on precision alone, making FAR the primary discriminating metric on those datasets in Sec.~\ref{sec:results-triplet}. 

\subsection{Precision, Coverage, and FAR Triplet}\label{sec:results-triplet}
Table~\ref{tab:operating_point} reports precision, coverage, FAR, and the aggregate score $\mathcal{S}$ averaged across VPR backbones. For uncertainty baselines we report results at two operating points: a label-free threshold (confidence score$\geq0.5$ after normalization), consistent with our deployment setting, and a threshold $\tau$ that maximizes the harmonic mean of precision and coverage using ground-truth labels.


The three metrics are interpreted jointly, since a verification method is useful only when it combines high precision and high coverage with low FAR. The aggregate score summarizes this trade-off in a single number and penalizes methods that optimize only one dimension, such as achieving high precision by rejecting nearly all queries (e.g., Qwen-7B on Nordland: $P = 0.99$, $C = 0.01$) or maintaining high coverage while accepting false positives (e.g., SuperPoint on Pitts250k: $C = 1.00$, $\text{FAR} = 1.00$). VLMs achieve the strongest overall trade-off, with Gemini and GPT-4.1 obtaining average $\mathcal{S}$ of 0.78 and 0.76. In contrast, uncertainty baselines are highly sensitive to threshold selection, with rankings changing substantially between the two thresholds, and their average $\mathcal{S}$ remains significantly lower than that of the VLMs under both strategies.

FAR reveals a substantial safety gap between conventional verification and VLM auditing. Under label-free thresholding, uncertainty methods often achieve low FAR only by severely reducing coverage (e.g., PA and SuperPoint accept no more than 30\% of queries), while under the best precision--coverage threshold these methods frequently maintain FAR between 0.7--1.0, accepting most incorrect retrievals. VLM verification requires no threshold selection and reduces FAR while maintaining practical precision and coverage. Gemini and GPT-4.1 achieve FAR below 0.30 across most datasets with high precision and coverage. 

These results suggest that estimating retrieval correctness using statistics derived from the same representation that produced the retrieval is fundamentally limited. Under both threshold-selection strategies the strongest conventional baseline is based on independent geometric reasoning (SuperPoint/SIFT), while semantic reasoning from VLMs provides the largest improvement.

\subsection{Risk-Coverage Analysis}\label{sec:results-riskcoverage}
Fig.~\ref{fig:risk_coverage} shows risk-coverage curves for all datasets, with uncertainty baselines swept across all thresholds and VLM operating points plotted as fixed markers. The curves reveal a structural limitation of conventional verification methods. Uncertainty estimates are concentrated toward the upper-right region of the plots, maintaining high coverage with high risk. Uncertainty metrics can reduce risk only by sacrificing large fractions of coverage, often degenerating toward near-zero acceptance before reaching meaningful risk reductions. This behavior is evident on AmsterTime, which has an unfiltered risk of 0.59. At the selected thresholds, uncertainty baselines accept 72--100\% of queries at risk 0.46--0.59, whereas VLMs achieve substantially lower risk at practical coverage levels. Across the remaining datasets VLMs consistently operate at lower risk at comparable coverage. On SanFran, Tokyo247, and StLucia, Gemini and GPT-4.1 achieve approximately 2--6$\times$ lower risk than the strongest conventional baselines. Pitts250k shows smaller differences because baseline retrieval performance is already high, but VLMs still provide lower FAR at comparable coverage. Uncertainty metrics offer a tunable trade-off, which allows selecting any point along the curve, whereas VLMs provide a fixed but generally superior operating point. 

\subsection{Limitations of AUC-PR}\label{sec:results-aucpr}
Fig.~\ref{fig:pr_curves} shows precision-recall curves for three datasets (AmsterTime, SanFran and Nordland) evaluated with EigenPlaces as the retrieval backbone. Curves for conventional verification methods are generated by thresholding the uncertainty scores, while VLM curves use the token-level probability of \textit{true} as a confidence score, making the two directly comparable. Across all three datasets, VLMs achieve competitive or superior AUC-PR. On AmsterTime, GPT-4.1 achieves an AUC-PR of 0.86, while the best verification baseline (PA-score) reaches 0.81. On SanFran and Nordland, both VLMs and baseline methods exceed 0.91, making their performance appear largely equivalent. This highlights the limitation of AUC-PR for the verification task, because it aggregates across every threshold while a deployed system operates at a single threshold, methods with similar AUC-PR can behave substantially differently once a threshold is selected. On Pitts250k, where baseline precision is already approximately 92\%, baseline verification methods achieve high AUC-PR despite FAR of 0.99--1.00, accepting nearly every false positive. Similarly, on Nordland, L2 and Qwen-7B achieve comparable AUC-PR despite fundamentally different behavior. L2 accepts almost all matches (FAR\,=\,0.93, coverage\,=\,0.94), whereas Qwen-7B rejects almost all matches (FAR\,=\,0.00, coverage\,=\,0.01). These results show that the precision-coverage-FAR triplet directly addresses AUC-PR's blind spots.

\begin{figure}[t]
\vspace{2mm}
\centering
\includegraphics[width=0.8\columnwidth]{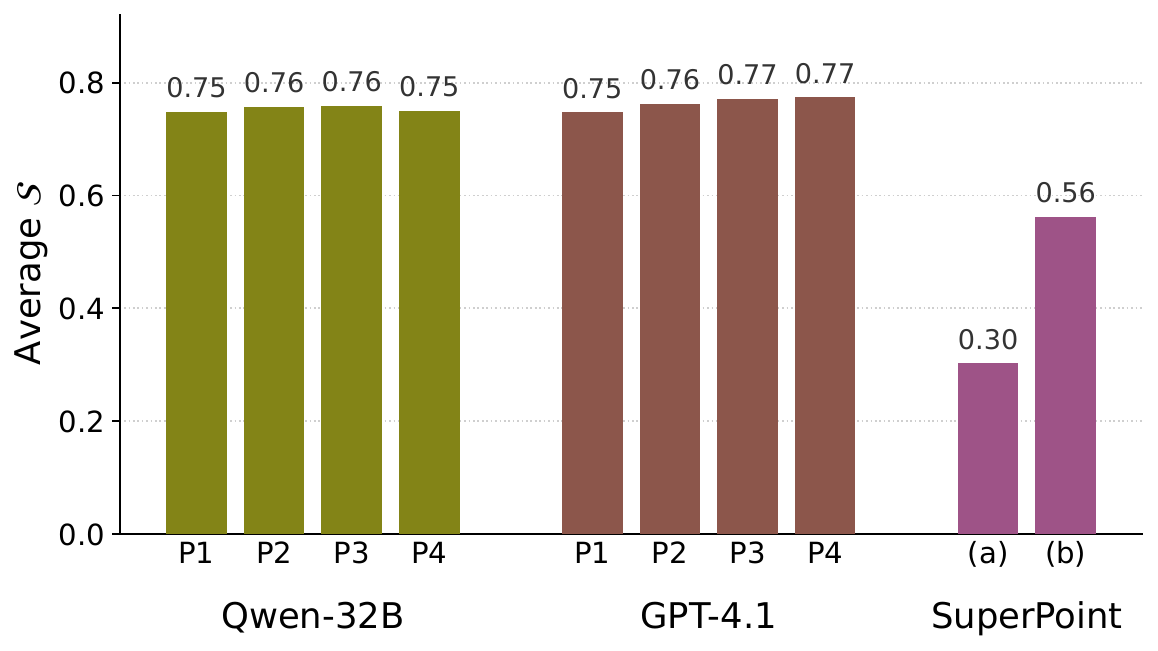}
\caption{Average aggregate score $\mathcal{S}$ for the four prompt formulations (P1--P4, Sec.~\ref{sec:experiment-prompting}) on Qwen-32B and GPT-4.1, with SuperPoint under both threshold-selection strategies. EigenPlaces backbone, averaged over AmsterTime, SanFran and Tokyo247. Prompt formulation changes $\mathcal{S}$ by at most 3 points, whereas SuperPoint changes by 26 points depending only on how its threshold is selected.}
\label{fig:prompt_sensitivity}
\end{figure}

\subsection{Prompt Sensitivity}\label{sec:prompt_sensitivity}
The auditing decision depends on a prompt rather than a numerical threshold, so we conduct a prompt sensitivity analysis. We compare the four prompt formulations described in Section~\ref{sec:experiment-prompting}, on three datasets using EigenPlaces as the retrieval backbone, for one open-weight and one closed-source model. Increasing strictness raises precision and reduces coverage and FAR, as expected. Precision varies by under 2 points across the four prompts, while coverage and FAR vary up to 6. As shown in Fig.~\ref{fig:prompt_sensitivity}, the aggregate score varies by at most 3 points across the full range of strictness and rewording, whereas SuperPoint differs by 26 points on the same datasets depending only on how its threshold is selected. Almost every prompt we evaluate outperforms SuperPoint under either threshold-selection strategy.


\subsection{Discussion}
Our failure analysis indicates that VLMs improve verification without introducing fundamentally new failure modes. Across all experiments, 99\% of VLM false-positive acceptances occur on image pairs that are also incorrectly accepted by conventional verification methods, so the matches VLMs fail to reject are almost entirely those that remain difficult for embedding- and geometry-based approaches. The dominant remaining failure mode is therefore reduced coverage. The causes of false rejection differ across models. Gemini and GPT-4.1 typically reject image pairs due to structural differences introduced by viewpoint changes (e.g., differing numbers of visible railway tracks or poles), whereas Qwen frequently cites dynamic scene attributes that the prompt instructs it to ignore (e.g., snow cover or seasonal lighting changes), suggesting that its reduced coverage stems from weaker instruction following rather than visual reasoning limitations. VLM auditing has substantially higher latency than any conventional method (Table~\ref{tab:operating_point}) and is infeasible for real-time per-query verification. The cost is more readily accepted where verification is not time-critical, for applications such as offline loop closure, where higher precision in exchange for higher inference time may be acceptable. Our evaluation uses public benchmark datasets. Validation on real-world robotic platforms remains future work.

\section{CONCLUSION}\label{sec:conclusion}
We introduced VPR Auditing, a post-retrieval verification framework in which a VLM assesses the correctness of a retrieved match directly from the images. Our results demonstrate that VLM-based auditing rejects far more false positives than conventional verification methods, without thresholds calibrated on ground truth. Verification should be evaluated at the single threshold a deployed system will use, rather than aggregated over all possible thresholds.


\bibliographystyle{IEEEtran}
\bibliography{IEEEexample}

\newpage

\vfill

\end{document}